\documentclass[sigconf]{acmart}

\usepackage{ragged2e} 
\usepackage{booktabs,makecell, multirow, tabularx}

\AtBeginDocument{%
  \providecommand\BibTeX{{%
    \normalfont B\kern-0.5em{\scshape i\kern-0.25em b}\kern-0.8em\TeX}}}

\setcopyright{acmcopyright}
\copyrightyear{2023}
\acmYear{2023}
\acmDOI{XXXXXXX.XXXXXXX}

\acmConference[]{}

\renewcommand\footnotetextcopyrightpermission[1]{}

\begin{document}

\title{FS-Depth: Focal-and-Scale Depth Estimation from a Single Image in Unseen Indoor Scene}


\author[author1]{Chengrui Wei, Meng Yang, Lei He, Nanning Zheng}
\renewcommand{\shortauthors}{ }

\begin{abstract}
It has long been an ill-posed problem to predict absolute depth maps from single images in real (unseen) indoor scenes. We observe that it is essentially due to not only the scale-ambiguous problem but also the focal-ambiguous problem that decreases the generalization ability of monocular depth estimation. That is, images may be captured by cameras of different focal lengths in scenes of different scales. In this paper, we develop a focal-and-scale depth estimation model to well learn absolute depth maps from single images in unseen indoor scenes. First, a relative depth estimation network is adopted to learn relative depths from single images with diverse scales/semantics. Second, multi-scale features are generated by mapping a single focal length value to focal length features and concatenating them with intermediate features of different scales in relative depth estimation. Finally, relative depths and multi-scale features are jointly fed into an absolute depth estimation network. In addition, a new pipeline is developed to augment the diversity of focal lengths of public datasets, which are often captured with cameras of the same or similar focal lengths. Our model is trained on augmented NYUDv2 and tested on three unseen datasets. Our model considerably improves the generalization ability of depth estimation by 41\%/13\% (RMSE) with/without data augmentation compared with five recent SOTAs and well alleviates the deformation problem in 3D reconstruction. Notably, our model well maintains the accuracy of depth estimation on original NYUDv2.
\end{abstract}





\begin{teaserfigure}
  \centerline{\includegraphics[width=17cm]{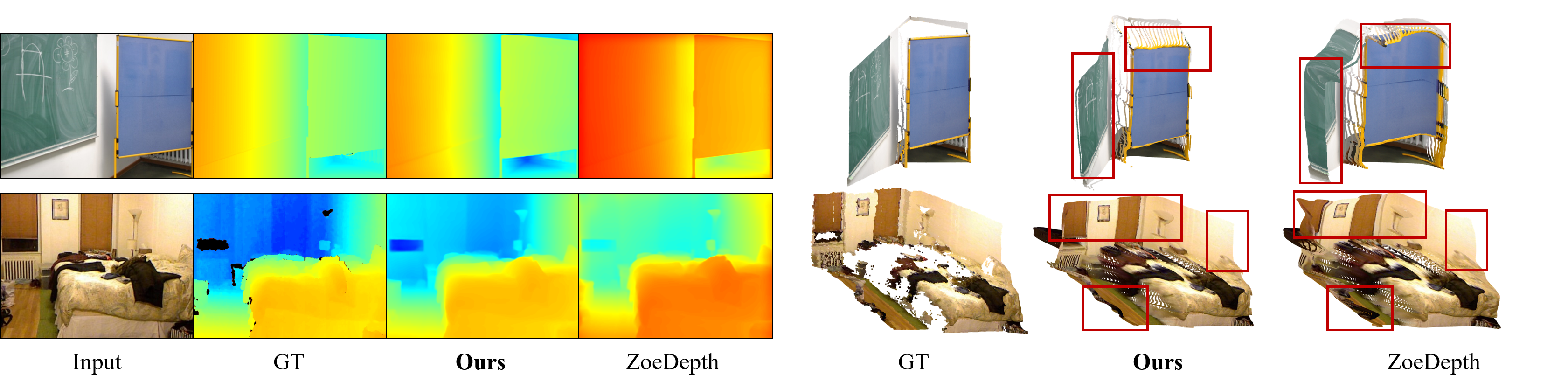}}
  \caption{ Depth map prediction from an image with different focal lengths and scene scales/scemantics. Left: Predicted depth maps, Right: depth maps in reconstructed 3D view. Our model accurately predicts absolute depth values in unseen scenes and well alleviates the serious deformation problem in 3D view (red boxes) compared with a recent SOTA model ZoeDepth \cite{2}.}
  \Description{Enjoying the baseball game from the third-base
  seats. Ichiro Suzuki preparing to bat.}
  \label{fig:teaser}
\end{teaserfigure}


\maketitle

\section{Introduction}
Monocular depth estimation is a fundamental task in the field of multimedia systems and robotics in indoor scenes. This task predicts pixel-by-pixel depth values as well as 3D geometry of the scene from a single 2D image. It well facilitates subsequent tasks of 3D scene perception such as human-robot interaction \cite{49,56}, augmented reality \cite{3,57}, 3D reconstruction \cite{50,55}, autonomous robots \cite{4,58}, and CAD model generation \cite{5}.

Monocular depth estimation is essentially ill-posed due to the scale-ambiguous problem of single images in real (unseen) scene. That is, a 2D image may correspond to infinite depth maps of different scales in 3D scene. In recent years, this task has made great advances thanks to the development of deep neural networks, which learns prior knowledge from RGB-Depth datasets and then regresses depth maps from single images. Recent advances can be classified into two branches including absolute depth estimation \cite{6,7,8,9,10} and relative depth estimation \cite{9,11,48,21}. Absolute depth estimation directly regresses the pixel-level distances (in meters) between the scene and the camera. Absolute depth estimation models are usually trained and tested on the dataset of a single scene to avoid the scale-ambiguous problem. Mixing multiple datasets with different scene scales may disturb the training of the network, and thereby significantly reduce the accuracy of depth estimation. As a result, most existing models tend to be overfitting in a single dataset and cannot be well generalized to unseen scenes. To address this problem, relative depth estimation \cite{11,21,48,28} learns the relationships (i.e. far or near) of points in the scene regardless of the scale-ambiguous problem. Therefore, relative depth estimation models can be well trained on multiple datasets with different scene scales and have excellent generalization ability in unseen scenes. However, relative depth maps are often limited in practical applications due to the lack of object distances compared with absolute depth maps. Recently, researchers attempted to combine relative/absolute depth estimation models \cite{2,12,13,48}, which are realized by first predicting relative depth maps from mixed datasets of different scene scales and then regressing absolute depth maps with a scale network. Therefore, these models could well predict absolute depth maps from single images and can be well generalized to unseen scene with different scales.

In this paper, we observe that it is essentially due to not only the scale-ambiguous problem of the scenes but also the focal-ambiguous of cameras to disturb the training of monocular depth estimation in unseen indoor scenes. That is, high-quality images may be captured by cameras of different focal lengths in scenes of different scales/semantics. It is acknowledged that cameras generally focus 3D scenes on CCD/CMOS \cite{53,54} planes by intrinsic optical imaging system \cite{45}. As a result, the focal length of cameras may vary in three cases in unseen scenes. First, the focal length of cameras may be changed smaller or larger in the scenes with different scales. This case corresponds to the common scale-ambiguous problem, which has been well studied in the literature \cite{46}. Second, the focal length of cameras may be changed smaller or larger in the scenes with different semantics such as tiny and huge objects. Third, the focal lengths of different camera configurations naturally differ with each other. The diversity of focal lengths may disturb the training of monocular depth estimation together with the diversity of scene scales, thereby reduce the accuracy of predicted depth maps in unseen scenes. Unfortunately, such focal-ambiguous problem of monocular depth estimation was not investigated sufficiently in the literature \cite{1,2}. Though the effect of scene scales on focal lengths has been partly avoided in joint relative/absolute depth prediction \cite{2,12}, focal lengths of cameras may still be changed in the other two cases in unseen scenes. Existing absolute depth estimation models are often trained/tested on the same dataset such as NYUDv2 \cite{20} and these models often achieve desire results in similar scenes. One reason lies that all images in these datasets are captured with cameras of the same or similar focal lengths. Therefore, these models cannot be well generalized to unseen indoor scenes due to not only the scale-ambiguous problem but also the focal-ambiguous problem \cite{1}.

In this paper, we further investigate the focal-ambiguous problem of monocular depth estimation in unseen indoor scenes. To this end, we develop a focal-and-scale depth estimation model, namely \textbf{FS-Depth}, to well learn absolute depth maps from single images with diverse focal lengths and scales in unseen scenes. First, we adopt a relative depth estimation network \cite{11} to learn relative depth maps from single images with diverse scales/semantics. This model is well pre-trained on multiple datasets of different scene scales/semantics. Second, we generate multi-scale features by mapping a single focal length to focal length features and concatenating them with intermediate features of different scales in relative depth estimation. Third, relative depth maps and multi-scale features are jointly fed into an absolute depth estimation network to predict accurate depth maps from single images. The recent \cite{2} is adopted as our base model, which well combined relative/absolute depth estimation and achieved state-of-the-art (SOTA) results. Finally, we well fine-tune our FS-Depth model by exploring a critical issue of learning rates that may be significantly disrupted by diverse focal lengths. In addition, we develop a new data augmentation pipeline to generate images with diverse focal lengths for training and testing, considering that most public datasets are captured with cameras of the same or similar focal lengths.

Our model is trained on NYUDv2, evaluated on three unseen datasets including iBims-1 \cite{14}, DIODE-Indoor \cite{15}, DIML-Indoor \cite{16}, and compared with five recent SOTAs \cite{2,6,7,10,38}. The results demonstrate that our model considerably improves the generalization ability of monocular depth estimation by 41\%/13\% (RMSE) on the three unseen datasets with/without data augmentation. 
Visual results demonstrate that our model consistently produces accurate depth values and alleviates the serious deformation problem in 3D reconstruction compared with SOTAs such as the examples in Figure  \ref{fig:teaser}.
One reason lies that our model well incorporates focal lengths in monocular depth estimation while focal length cannot be directly fed into existing models. Notably, our model well maintains the accuracy of monocular depth estimation on the test dataset of NYUDv2, which has the same focal length and scene scales with the training dataset. 

The main contributions are summarized as follows.
\begin{itemize}
\item  	We observe that it is essentially due to not only the scale-ambiguous problem but also the focal-ambiguous problem that decreases the generalization ability of monocular depth estimation in unseen indoor scenes.
\item 	We develop a focal-and-scale depth estimation model to well learn absolute depth maps from single images with diverse focal lengths and scene scales/semantics. Our model considerably improves the generalization ability on unseen datasets compared with recent SOTAs.
\item We develop a new data augmentation pipeline to collect RGB-Depth datasets with diverse focal lengths, considering that most public datasets are captured with cameras of the same or similar focal lengths.
\end{itemize}

\section{Related Work}
\subsection{Absolute depth estimation}
Absolute depth estimation regresses pixel-level distances between the scene and the camera. Absolute depth estimation has been widely investigated in the past decade \cite{51,6,7,8,9,10,39}. For example,  Eigen \textit{et al.} \cite{39} first proposed a coarse-fine network to obtain the depth with the constraint of a scale-invariant loss.
In addition, some researchers improved the accuracy of absolute depth estimation by incorporating prior knowledges. For example, He \textit{et al.} \cite{1} first learned absolute depth maps from single images with deep neural network embedding focal length. Saxena \textit{et al.} \cite{52} transformed 3D coordinates into planar parameters based on geometric prior, and utilized mean pane loss to improve the prediction of depth values. Most existing models are trained and tested on a single dataset such as NYUDv2. Training on multiple datasets may degrade the accuracy of these models due to the scale-ambiguous problem. As a result, most existing models tend to be overfitting on a single dataset and the generalization ability was not well investigated. Therefore, most existing models cannot be well generalized to unseen scenes. The reason lies that images may be captured by cameras of different focal lengths in scenes of different scales/semantics in unseen scenes. In this paper, we well investigate the generalization ability of absolute depth estimation in unseen scenes.

\subsection{Relative depth estimation}
Relative depth estimation \cite{9,11,13,21} learns the relationships of points in the scene instead of absolute distances. This task was first  investigated in \cite{48}, which also introduced a new dataset consisting of images annotated with relative depth values. Recently, Ranftl \textit{et al.} \cite{11} proposed a robust training objective that is invariant to changes in depth range and scale. Relative depth estimation well handles the scale-ambiguous problem by ignoring the scales of different scenes. Therefore, it can be well trained on multiple datasets and can be easily generalized to unseen scenes. Relative depth estimation could be well used in a few applications such as human-robot interaction. However, these models are often limited in other scenarios due to the lack of object distances. For example, it may require absolute depth values in 3D reconstruction and autonomous robots. In this paper, we aim to directly estimate absolute depth values in unseen scenes.

\subsection{Joint relative/absolute depth estimation}
Recently, researchers combined relative/absolute depth estimation to well learn absolute depth values from single images \cite{48, 2, 12}. For example, Jun \textit{et al.} \cite{12} decomposed the relative depth into normalized depth and scale features, and proposed a multi-code network in which the absolute depth decoder utilized relative depth features from the gradient and normalized depth decoders. Benefiting from relative depth estimation, these models can be well trained on multiple datasets and thereby are well generalized to unseen scenes. In this paper, we adopt a recent model of joint relative/absolute depth estimation \cite{2} as base model to handle the scale-ambiguous problem. However, we observe that it is essentially due to not only the scale-ambiguous problem of the scenes but also the focal-ambiguous of cameras that degreases the generalization ability of monocular depth estimation. Therefore, we additionally take focal length as input of the base model to well handle the focal-ambiguous problem.

\begin{figure*}
  \centerline{\includegraphics[width=16cm]{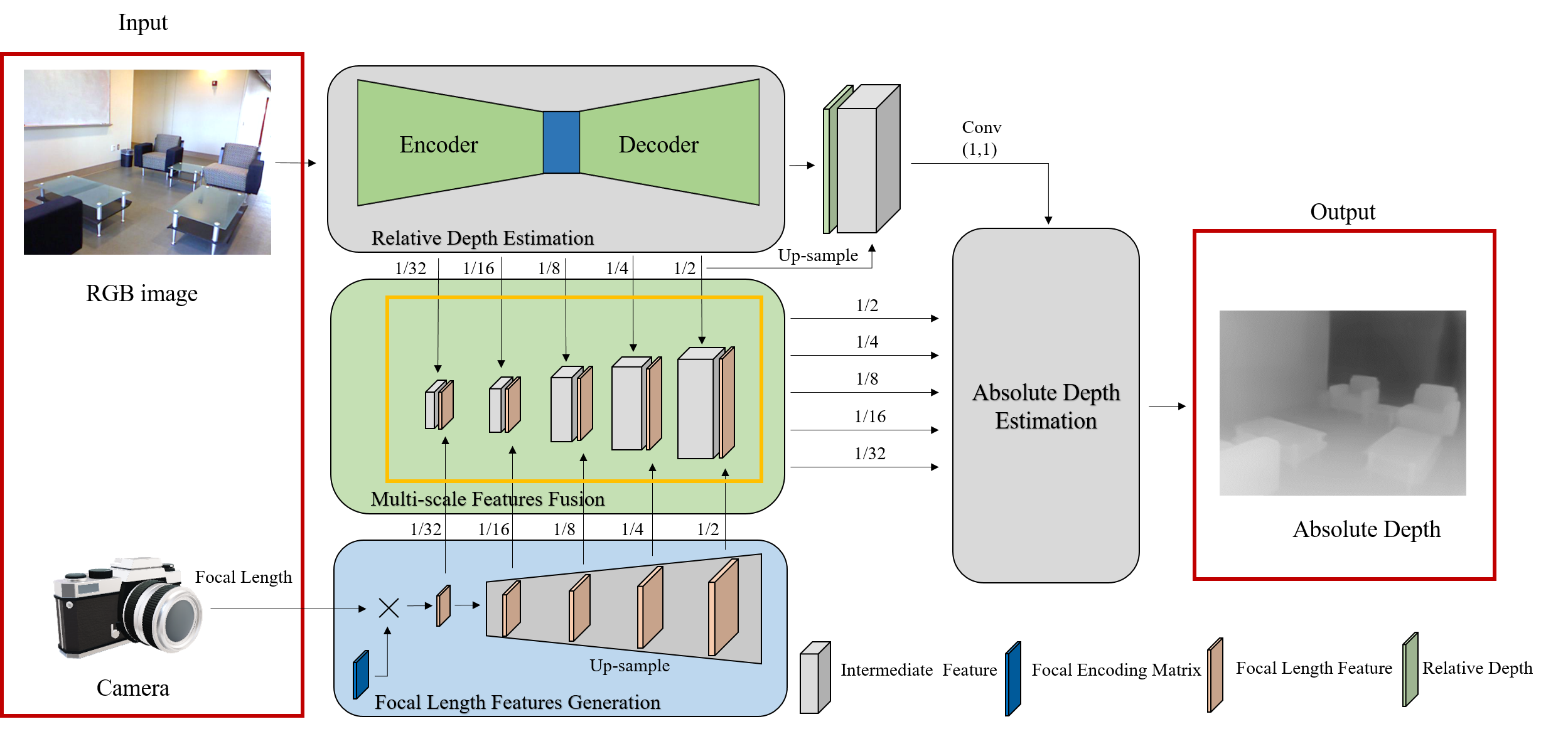}}
  \caption{ Our FS-Depth model comprises four main components including relative depth estimation, focal length features generation, multi-scale features fusion, and absolute depth estimation. The framework is described in Section 3.1. }  \label{fig:net}
  \Description{Enjoying the baseball game from the third-base
  seats. Ichiro Suzuki preparing to bat.}
\end{figure*}


\section{Method}
\subsection{Framework}
Figure \ref{fig:net} shows the network architecture of our FS-Depth model. It takes a single RGB image and the focal length of camera as the input and output an absolute depth map. First, a relative depth estimation network is used to learn relative depths from single images with diverse scales/semantics (in \textbf{Section 3.2}). Second, focal length features are generated by multiplying a single focal length to a learnable focal encoding matrix and up-sampling it to multiple resolutions (in \textbf{Section 3.3}). Third, multi-scale features are generated by concatenating focal length features of different resolutions with intermediate features of different scales in relative depth estimation (in \textbf{Section 3.4}). Finally, relative depth maps and multi-scale features are jointly fed into an absolute depth estimation network to predict accurate depth maps (in \textbf{Section 3.5}). Our model is well fine-tuned by exploring a critical issue of learning rates (in \textbf{Section 3.6}). The recent \cite{2} is adopted as our base model, which well combined relative/absolute depth estimation networks and utilized a scale-invariant logarithmic loss \cite{39}. In addition, we develop a new data augmentation method to collect datasets with diverse focal lengths for training and testing (in \textbf{Section 5}). Our FS-Depth model is introduced in detail in the following.

\subsection{Relative depth estimation}
Relative depth estimation is used to learn relative depths, \textit{i.e.} the relationships (far or near) of pixels, from single images. It ignores the scales of different scenes, therefore, can be well trained on mixed datasets. 
There are many relative depth estimation models such as \cite{48,11,21}, among which the recent MiDaS \cite{11} is the most famous one. This model was realized based on the backbone ResNeXt-101-WSL \cite{resnext} and a normalized scale loss function. It was pretrained on ImageNet \cite{imagenet} and then further trained on mixing multiple datasets with a pareto-optimal strategy to well learn relative depth maps from single images. This model was adopted in our base model ZoeDepth \cite{2}. Notably, it used an updated version with the DPT encoder-decoder architecture \cite{21} and a transformer-based backbone BEiT384-L \cite{47} instead to further improve the accuracy of relative depth estimation. In our solution, this well-trained model in \cite{2} is directly adopted to predict high-quality relative depth maps from single images of diverse scene scales/semantics.

\subsection{Focal length features generation}
Our FS-Depth model jointly takes a single RGB image $\mathbf{x}$ and reference focal length of the camera $f$ as inputs during training and testing. Unfortunately, a single value of focal length $f$ cannot be directly fed into the network together with the vector $\mathbf{x}$ of RGB image. One reason lies that typical backbone networks with pre-training weights are often designed for single RGB images, which are only applicable to inputs with three channels. Therefore, no pre-training weights are available to handle additional input of focal lengths. We attempt to input the focal length in the middle stage of the network, rather than at the beginning along with the RGB image. Specifically, we directly embed the focal length into the absolute depth estimation network. To this end, we introduce a learnable focal encoding matrix $\mathbf{M}$ inspired by the position encoding design in ViT \cite{26}. In order to align with the minimal intermediate feature maps of input RGB image, we set the size of the matrix to $12\times16$. We then generate a focal length feature for the network by multiplying the focal length value with the matrix as $\mathbf{F}=f\times\mathbf{M}$. The focal length feature is up-sampled to obtain focal length features of different scales by a linear interpolation operation. In our model, five focal length features $\mathbf{F}_{j}  \ (1\leq j \leq 5)$ of the scales 1/2, 1/4, 1/8, 1/16, and 1/32 are generated, where $j$ denotes the index of the five scales.

\subsection{Multi-scale features fusion}
The generated focal length features of different scales $\mathbf{F}_{j} \  (1\leq j \leq 5)$ are then embedded into the absolute depth estimation network. To this end, we similarly extract five intermediate features $\mathbf{N}_{j}$ of different scales 1/2, 1/4, 1/8, 1/16, and 1/32 from the relative depth estimation network, where $j$ denotes the index of the intermediate layers of the image $\mathbf{x}$. Specifically, the five intermediate features are from the bottleneck of the encoder-decoder architecture and the four subsequent levels of the decoder in MiDaS. The focal length features $\mathbf{F}_{j}$ are then concatenated with the intermediate features $\mathbf{N}_{j}$ at different scales, resulting in our final multi-scale features $\mathbf{feature}_{j}=[\mathbf{N}_{j},\mathbf{F}_{j}]$.  Our multi-scale features add the focal length value to intermediate features of different scales in relative depth estimation, which can well consider the influence of focal length on depth estimation more comprehensively. Once the focal length is added to a single layer of the network, the influence of focal length on subsequent layers may fade gradually. Our final multi-scale features are then fed into the subsequent absolute depth estimation network. The effect of the multi-scale features will be verified in the ablation study (in \textbf{Section 5.6}).

\subsection{Absolute depth estimation}
The absolute depth estimation network jointly takes our multi-scale features and output of relative depth estimation network to predict accurate depth maps. In our base model, a tensor is first obtained by concatenating the relative depth map with the last intermediate feature in relative depth estimation \cite{2}. Then the model predicts discrete depth bins \cite{22, 23, 24, 25} from intermediate features of relative depth estimation and the probability of each pixel corresponding to different bins from the tensor. Finally, the model obtains absolute depth value of each point by weighted summation of discrete depth bins based on the probability distribution. In our solution, we only adjust the number of channels in depth bin prediction to well accommodate our multi-scale features with focal length.

\subsection{Learning rates determination}
Our FS-Depth model comprises three networks including the relative depth estimation network , the focal length features generation network and the absolute depth estimation network in Figure \ref{fig:net}. However, only  the focal length features generation network and the absolute depth estimation network have access to the focal length and the relative depth estimation network is well pre-trained in advance. Therefore, learning rates of the three networks play an important role in our solution. If the learning rates of the three networks are not properly constrained, errors induced by focal length may disturb the well-trained relative depth estimation network and thereby reduce the performance of our model. In our solution, we determine the learning rate of the relative depth estimation network to be $1/50$ of  the focal length features generation network and the absolute depth estimation network. The effect of the learning rates will be verified in the ablation study (in \textbf{Section 5.6}).

\begin{figure}[h]
  \centering
  \includegraphics[width=6cm]{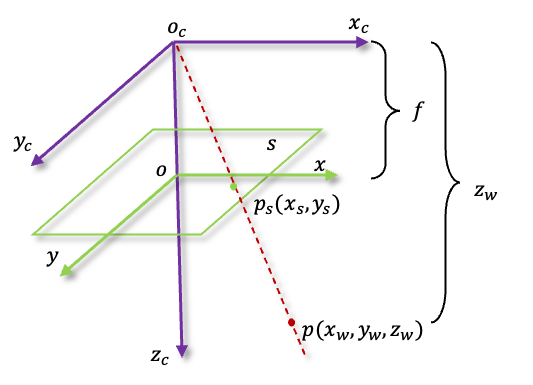}
  \caption{Principle of data augmentation to generate RGB-Depth pairs with diverse focal lengths.}  \label{chengxiang}
  \Description{A woman and a girl in white dresses sit in an open car.}
\end{figure}

\section{RGB-Depth Data Augmentation}
Our FS-Depth model requires RGB-Depth datasets with diverse focal lengths and scene scales for training and testing. The scene scales generally vary across public datasets. However, most public datasets are collected with cameras of the same or similar focal lengths such as NYUDv2. In this paper, we propose a new data augmentation pipeline to generate datasets with diverse focal lengths from public datasets above.
Figure \ref{chengxiang} illustrates the basic principle of the camera to capture an image, where $o_{c}$ denotes the position of the camera, $p=(x_w, y_w, z_w)$ denotes the coordinates of a point in the 3D scene, $s$ denotes the image plane of the camera, $p_s=(x_s, y_s)$ denotes the projected point of the point $p$ in the image plane, and $f$ denotes the focal length of the camera. It follows that:
\begin{equation}
\frac{x_{w}}{x_{s}}=\frac{y_{w}}{y_{s}}=\frac{z_{w}}{f}.\label{bili} \tag{1}
\end{equation}

When the coordinates $p_s=(x_s, y_s)$ and the focal length $f$ are simultaneously scaled by a factor $k$, the equation (\ref{bili}) still holds. The coordinates of point $p$ in the real scene remains unchanged. Therefore, by scaling the coordinates of each pixel in the RGB image by a factor $k$, the resulting RGB-Depth pair is equivalent to the one captured by a camera with the focal length  $kf$.

Similarly, when the coordinates $p_s=(x_s, y_s)$ are scaled by a factor $k$ while the focal length  $f$ is kept unchanged, the depth value $z_w$ needs to be scaled by $1/k$ to maintain the equation (\ref{bili}). Therefore, by scaling the coordinates of all pixels in the RGB image by a factor $k$ and scaling the values of all pixels in the depth map by a factor $1/k$, the resulting RGB-Depth pair is equivalent to the one captured by a camera with unchanged focal length $f$.

Based on the above analysis, the data augmentation pipeline is described as follows.

$\mathbf{Step \ 1}.$ Given a random factor $k \ (0.7\leq k\leq 1)$ and an RGB-Depth pair of size $m\times n$ that is captured with the focal length $f$, we crop a new RGB-Depth pair of small size $km\times kn$ at the center of the image.

$\mathbf{Step \ 2}.$ Up-sample the new RGB-Depth pair of size $km\times kn$ to the size of $m\times n$ by nearest-neighbor interpolation. The generated RGB-Depth pair is equivalent to the one captured by a camera with the focal length $f/k$. Notably, pixel values in the depth map are not changed in this step.

$\mathbf{Step \ 3}.$ Up-sample the new RGB-Depth pair of the size $km\times kn$ to the size of $m\times n$ by nearest-neighbor interpolation and rescale the depth values $z_w$ by a factor $k$ to obtain a depth map with the values $kz_w$. The generated RGB-Depth pair is equivalent to the one captured by a camera with the focal length $f$. Notably, pixel values in the depth map are changed in this step.

\begin{figure}[h]
  \centering
  \includegraphics[width=\linewidth]{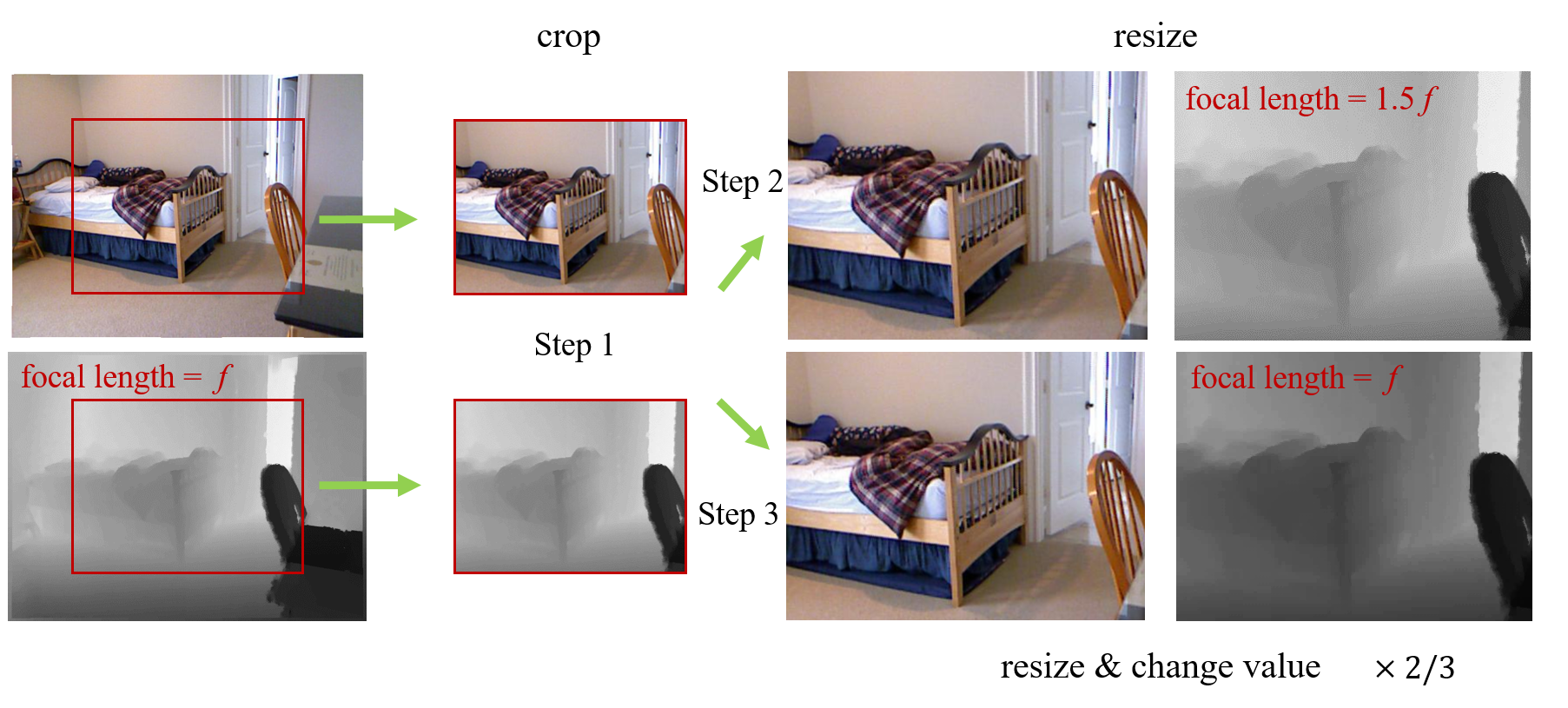}
  \caption{An example of our data augmentation pipeline.}  \label{step}
  \Description{A woman and a girl in white dresses sit in an open car.}
\end{figure}

Figure \ref{step} shows an example of our data augmentation pipeline. Notably both Step 2 and Step 3 in our data augmentation pipeline are used to generate RGB-Depth datasets with diverse focal lengths. When only Step 2 is activated, multiple RGB images with different focal lengths correspond to the same depth range. The network may be overfitting to a single depth range and fail to well learn the meaning of input focal lengths. Thereby, it may significantly reduce the generalization ability of the network. When only Step 3 is activated, some prior knowledge of the real scene such as the size of objects may be destroyed due to the change of depth values. The monocular depth estimation network could not well learn such prior knowledge in the real scene and thereby its generalization ability may be decreased. In our solution, the RGB-Depth pairs in Step 2 and Step 3 are mixed in a ratio of $60\%:40\%$ to generate an augmented training dataset with diverse focal lengths. The effectiveness of our data augmentation pipeline will be verified in the ablation study (in \textbf{Section 5.6})

\section{Experiments and Analysis}
\subsection{Experiment setting}
$\mathbf{Datasets}.$ Our model is trained on the training dataset of NYUDv2. The generalization ability of our FS-Depth model is tested on three unseen datasets iBims-1 \cite{14}, DIODE-Indoor \cite{15}, and DIML-Indoor \cite{16} with different scene scales/semantics. The dataset NYUDv2 consists 464 indoor scenes. We use the official split which consists 24231 examples from 249 scenes provided by previous methods \cite{38} for training and the test set consists 654 examples from 215 scenes based on \cite{39}. The dataset iBims-1 contains 100 test examples. The DIODE-Indoor dataset contains 325 test examples. The DIML-Indoor dataset contains 503 test examples.

\noindent $\mathbf{Train}.$ We conduct the experiments on two NVIDIA GeForce RTX 3090 graphics cards with a batch size of 4. We employ the AdamW \cite{adamw} optimizer with a weight decay of 0.01. The learning rates of the focal length features generation network and  the absolute depth estimation network is set to $1.6^{-4} $, while the learning rate of the relative depth estimation network is set to $1/50 \times 1.6^{-4} $. We train our model for 5 epochs on the training dataset of NYUDv2. Notably the relative depth estimation network in our model has been pre-trained on twelve datasets following \cite{11,2}.

\subsection{Results on augmented datasets (unseen)}

\begin{table*}[]
  \caption{Generalization ability on augmented datasets (unseen). Best results are in bold, second best are underlined. Our model consistently achieves a considerable improvement on these unseen datasets compared with the second-best model.}
  \label{fanhua_a}
  \setlength{\tabcolsep}{1.5mm}{
\begin{tabular}{c|ccc|ccc|ccc}
\hline
\multirow{2}{*}{Method}                              & \multicolumn{3}{c|}{iBims-1}                                 & \multicolumn{3}{c|}{DIODE-Indoor}                                                                                         & \multicolumn{3}{c}{DIML-Indoor}                                                                                          \\
                                                     & $\delta_{1} \uparrow$ & $REL \downarrow$ & $RMSE \downarrow$ & \multicolumn{1}{l}{$\delta_{1} \uparrow$} & \multicolumn{1}{l}{$REL \downarrow$} & \multicolumn{1}{l|}{$RMSE \downarrow$} & \multicolumn{1}{l}{$\delta_{1} \uparrow$} & \multicolumn{1}{l}{$REL \downarrow$} & \multicolumn{1}{l}{$RMSE \downarrow$} \\ \hline
BTS \cite{38} & 0.252                 & 0.288            & 1.450             & 0.149                                     & 0.468                                & 2.232                                  & 0.183                                     & 0.375                                & 1.289                                 \\
AdaBins \cite{6}                     & 0.159                 & 0.374            & 1.509             & 0.101                                     & 0.541                                & 2.341                                  & 0.167                                     & 0.366                                & 1.240                                 \\
LocalBins \cite{7}                   & 0.237                 & 0.330            & 1.235             & 0.132                                     & 0.497                                & 2.230                                  & 0.259                                     & 0.314                                & 1.124                                 \\
NeWCRFs \cite{10}                    & 0.219                 & 0.323            & 1.277             & 0.121                                     & 0.526                                & 2.150                                  & 0.282                                     & 0.296                                & 1.007                                 \\
ZoeDepth \cite{2}                    & \underline{0.280}           & \underline{0.274}      & \underline{1.069}       & \underline{0.232}                               & \underline{0.393}                          & \underline{1.836}                            & \underline{0.297}                               & \underline{0.265}                          & \underline{0.833}                           \\ \hline
\textbf{Ours}                                        & \textbf{0.741}        & \textbf{0.169}   & \textbf{0.574}    & \textbf{0.508}                            & \textbf{0.310}                       & \textbf{1.327}                         & \textbf{0.797}                            & \textbf{0.151}                       & \textbf{0.430}                        \\
Improvement                                & 164.6\%               & 38.3\%           & 46.3\%            & 118.9\%                                   & 21.1\%                               & 27.7\%                                 & 168.4\%                                   & 43.0\%                               & 48.4\%                               \\ \hline
\end{tabular}}
\end{table*}

\begin{figure*}
  \centerline{\includegraphics[width=16cm]{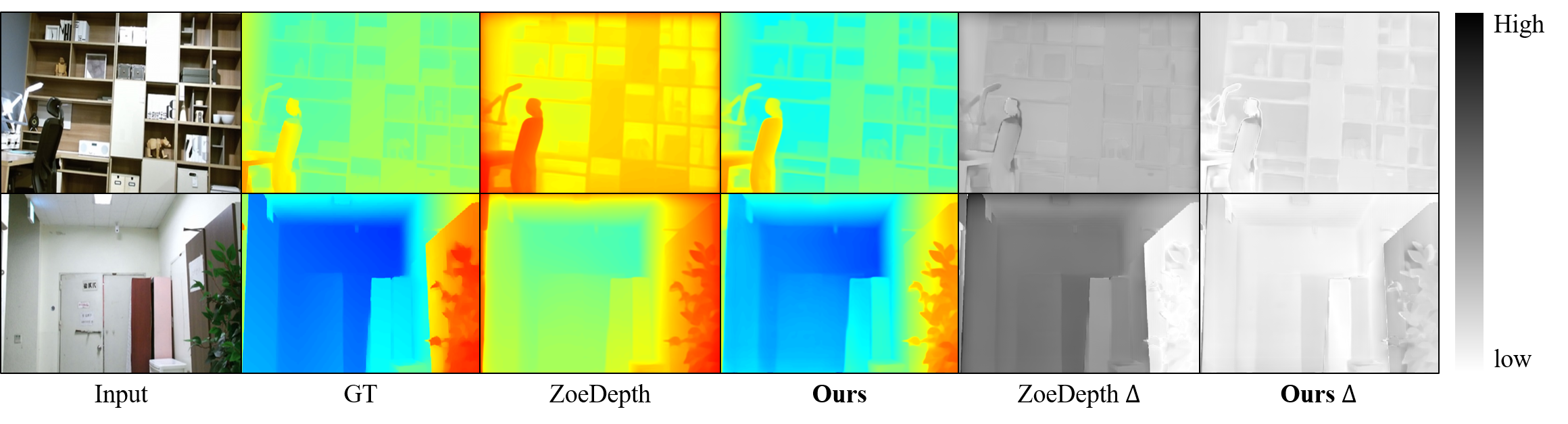}}
  \caption{Predicted depth maps on augmented DIML-Indoor (unseen). $\Delta$ denotes the error maps of predicted depth maps compared with GT. Our method consistently produces accurate depth maps with smaller (white) values in the error maps.}\label{diml_a}
  \Description{Enjoying the baseball game from the third-base
  seats. Ichiro Suzuki preparing to bat.}
\end{figure*}

We first test the generalization ability of our model on the augmented datasets with diverse focal lengths and scene scales. Table \ref{fanhua_a} concludes the results of our model using three common metrics. Five recent monocular depth estimation models are used as the baselines, among which the two models NewCRFs \cite{10} and ZoeDepth \cite{2} achieve the SOTA performance. Notably both our model and the baseline models are not fine-tuned on these test datasets. Our model consistently achieves a significant improvement on all these datasets using different metrics. For example, RMSE of predicted depth maps are reduced by 46.3\%, 27.7\%, and 48.4\%, respectively, on the three datasets compared with the second-best model. It indicates that our model is well generalized to the datasets that are collected by cameras of different focal lengths in unseen scenes. The reason lies that incorporating focal length can effectively improve the adaptability of our model to the cameras with different focal lengths. By comparison, though some of the baselines are well generalized to unseen scenes of different scales, focal length cannot be directly incorporated in all these models.
Figure \ref{diml_a} shows the visual results of predicted depth maps on DIML-Indoor. Our model not only well predicts the structures of depth maps, but more importantly, estimate accurate absolute depth values. It can be clearly seen in Figure \ref{diml_a_3d} that our model well alleviates the deformation problem in 3D reconstruction (in red boxes) compared with the second-best model.

\begin{figure*}
  \centerline{\includegraphics[width=16cm]{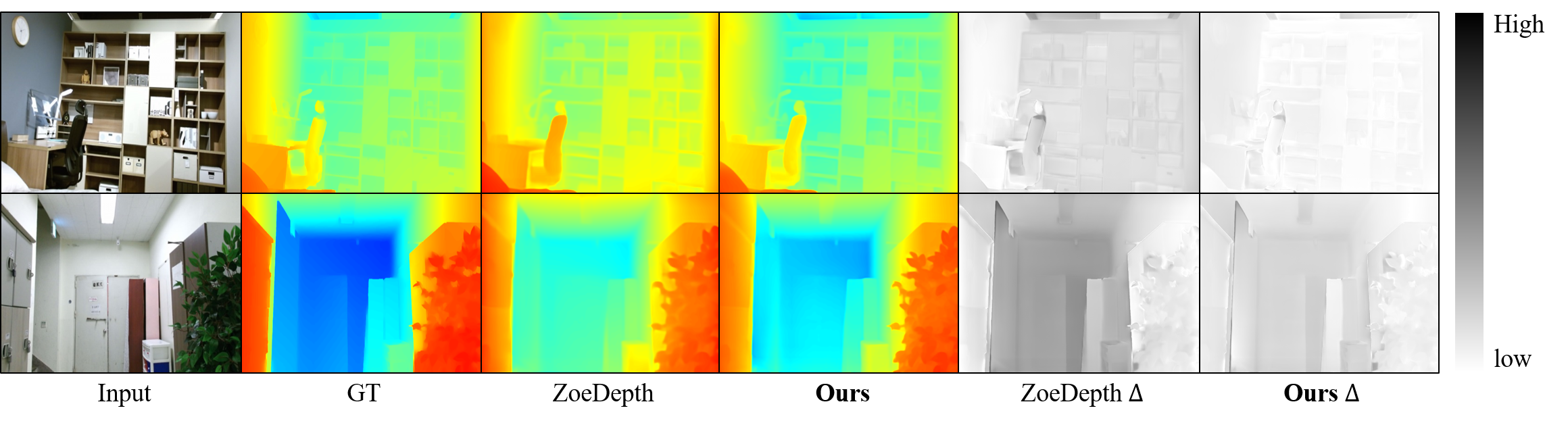}}
  \caption{Predicted depth maps on original Diml-Indoor (unseen). }\label{diml_o}
  \Description{Enjoying the baseball game from the third-base
  seats. Ichiro Suzuki preparing to bat.}
\end{figure*}


\subsection{Results on original datasets (unseen)}
\begin{figure}[h]
  \centering
  \includegraphics[width=8cm]{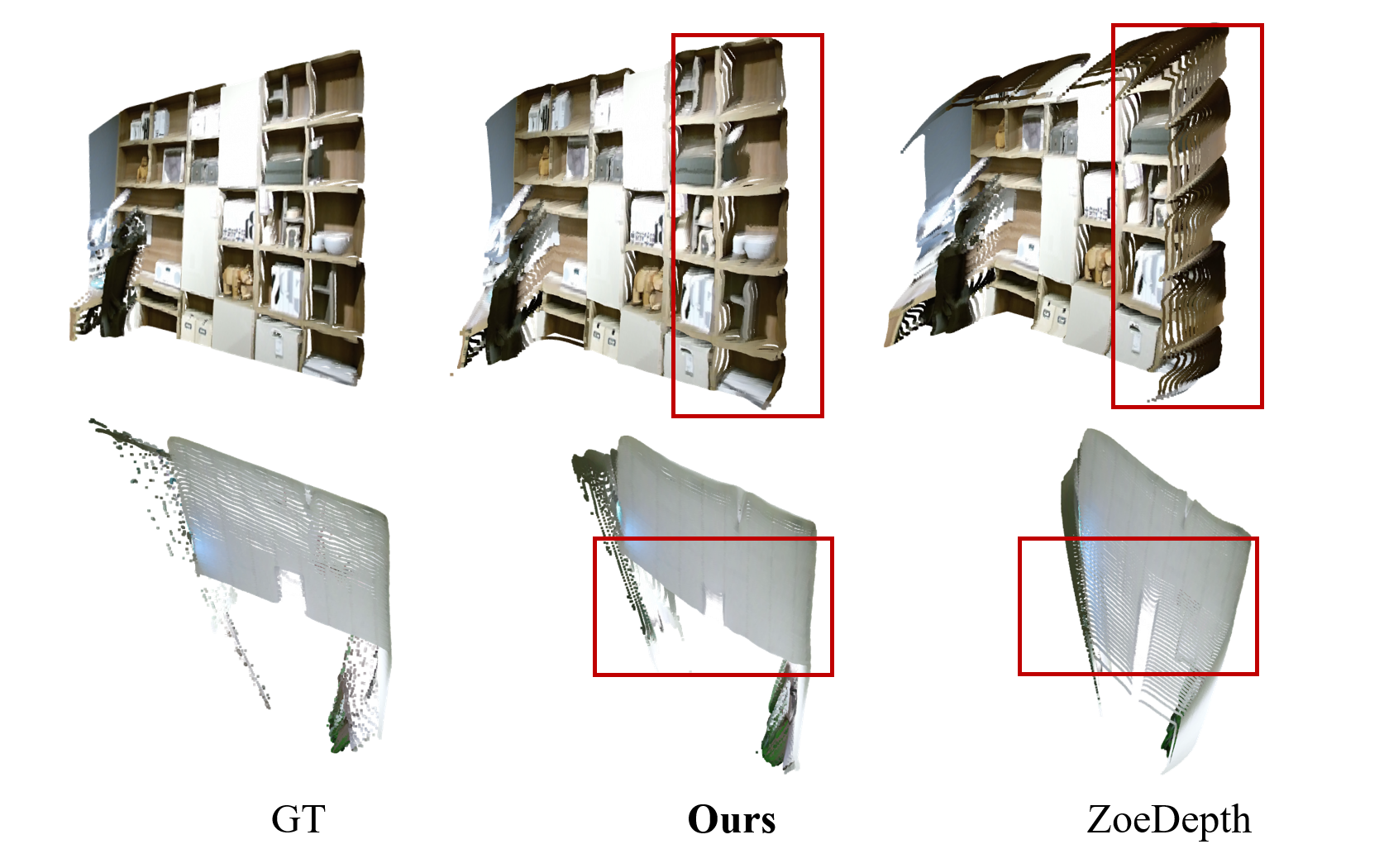}
  \caption{Reconstructed 3D scenes on augmented DIML-Indoor (unseen). Our model achieves the similar results compared with GT while the second-best model encounters the deformation problem in the red box.}  \label{diml_a_3d}
  \Description{A woman and a girl in white dresses sit in an open car.}
\end{figure}

\begin{figure}[h]
  \centering
  \includegraphics[width=8cm]{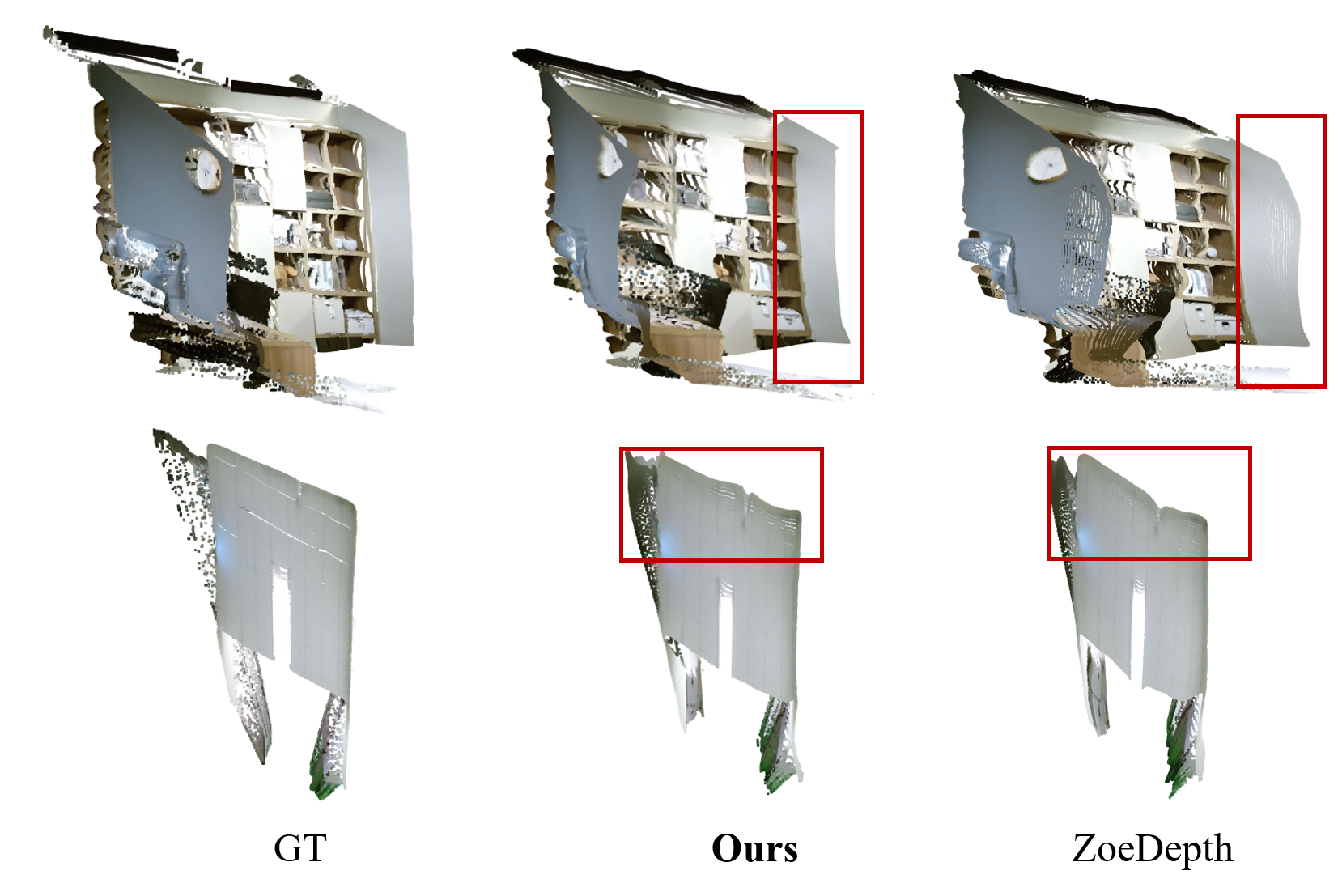}
  \caption{Reconstructed 3D scenes on augmented DIML-Indoor (unseen).}  \label{diml_o_3d}
  \Description{A woman and a girl in white dresses sit in an open car.}
\end{figure}

We then evaluate the generalization ability of our model on the original datasets of iBims-1, DIODE-Indoor, and DIML-Indoor. Notably the focal lengths across the test datasets and the training dataset NYUDv2 are still different, though each dataset is collected by cameras with the same or similar focal lengths.  Similarly, the five recent models are used as baselines. Table \ref{fanhua_o} concludes the results of our model and the baselines. Our model consistently outperforms all the baselines with different metrics. For example, RMSE of predicted depth maps are reduced by 17.3\%, 14.2, and 18.8\%, respectively, on the three datasets compared with the second-best model. In indicates that our model still works effectively on original public datasets. Figure \ref{diml_o} and \ref{diml_o_3d} show the visual results of predicted depth maps and reconstructed 3D scenes on a dataset. The results are similar to Figure \ref{diml_a} and \ref{diml_a_3d}.

\begin{table*}[]
  \caption{Generalization ability on original datasets (unseen).}
  \label{fanhua_o}
  \setlength{\tabcolsep}{1.5mm}{
\begin{tabular}{c|ccc|ccc|ccc}
\hline
\multirow{2}{*}{Method}                              & \multicolumn{3}{c|}{iBims-1}                                 & \multicolumn{3}{c|}{DIODE-Indoor}                                                                                         & \multicolumn{3}{c}{DIML-Indoor}                                                                                          \\
                                                     & $\delta_{1} \uparrow$ & $REL \downarrow$ & $RMSE \downarrow$ & \multicolumn{1}{l}{$\delta_{1} \uparrow$} & \multicolumn{1}{l}{$REL \downarrow$} & \multicolumn{1}{l|}{$RMSE \downarrow$} & \multicolumn{1}{l}{$\delta_{1} \uparrow$} & \multicolumn{1}{l}{$REL \downarrow$} & \multicolumn{1}{l}{$RMSE \downarrow$} \\ \hline
BTS \cite{38} & 0.538                 & 0.231            & 0.919             & 0.210                                     & 0.418                                & 1.905                                  & 0.652                                     & 0.223                                & 0.640                                 \\
AdaBins \cite{6}                     & 0.555                 & 0.212            & 0.901             & 0.174                                     & 0.443                                & 1.963                                  & 0.657                                     & 0.216                                & 0.623                                 \\
LocalBins \cite{7}                   & 0.558                 & 0.211            & 0.880             & 0.229                                     & 0.412                                & 1.853                                  & 0.689                                     & 0.208                                & 0.587                                 \\
NeWCRFs \cite{10}                    & 0.548                 & 0.206            & 0.861             & 0.187                                     & 0.404                                & 1.867                                  & 0.708                                     & 0.204                                & 0.553                                 \\
ZoeDepth \cite{2}                    & \underline{0.658}           & \underline{0.169}      & \underline{0.711}       & \underline{0.376}                               & \underline{0.327}                          & \underline{1.588}                            & \underline{0.646}                               & \underline{0.177}                          & \underline{0.536}                           \\ \hline
\textbf{Ours}                                        & \textbf{0.787}        & \textbf{0.145}   & \textbf{0.588}    & \textbf{0.522}                            & \textbf{0.283}                       & \textbf{1.363}                         & \textbf{0.796}                            & \textbf{0.147}                       & \textbf{0.414}                        \\
Improvement                                & 19.6\%               & 14.2\%           & 17.3\%            & 38.8\%                                   & 13.5\%                               & 14.2\%                                 & 23.2\%                                   & 16.9\%                               & 22.8\%                               \\ \hline
\end{tabular}}
\end{table*}

\begin{table*}[]
  \caption{Ablation study of focal length incorporation on three original datasets (unseen).}
  \label{cmp}
    \setlength{\tabcolsep}{1.5mm}{
\begin{tabular}{c|ccc|ccc|ccc}
\hline
\multirow{2}{*}{Method}                              & \multicolumn{3}{c|}{iBims-1 \cite{14}}                                 & \multicolumn{3}{c|}{DIODE-Indoor \cite{15}}                                                                                         & \multicolumn{3}{c}{DIML-Indoor \cite{16}}                                                                                          \\
                                                     & $\delta_{1} \uparrow$ & $REL \downarrow$ & $RMSE \downarrow$ & \multicolumn{1}{l}{$\delta_{1} \uparrow$} & \multicolumn{1}{l}{$REL \downarrow$} & \multicolumn{1}{l|}{$RMSE \downarrow$} & \multicolumn{1}{l}{$\delta_{1} \uparrow$} & \multicolumn{1}{l}{$REL \downarrow$} & \multicolumn{1}{l}{$RMSE \downarrow$} \\ \hline
base model & 0.658                 & 0.169            & 0.711             & 0.376                                     & 0.327                                & 1.588                                 & 0.646                                     & 0.177                                & 0.536                                 \\
single-scale & 0.738                 & 0.151            & 0.637             & 0.471                                     & 0.294                                & 1.428                                  & 0.767                                     & 0.150                                & 0.437                                 \\

\textbf{mutil-scale}                                        & \textbf{0.787}        & \textbf{0.145}   & \textbf{0.588}    & \textbf{0.522}                            & \textbf{0.283}                       & \textbf{1.363}                         & \textbf{0.796}                            & \textbf{0.147}                       & \textbf{0.414}                        \\
\hline
\end{tabular}}
\end{table*}

\begin{table*}[]
  \caption{Ablation study of the data augmentation ratio on three original datasets (unseen).}
  \label{ratio}
  \setlength{\tabcolsep}{1.75mm}{
\begin{tabular}{c|ccc|ccc|ccc}
\hline
\multirow{2}{*}{Ratio}                              & \multicolumn{3}{c|}{iBims-1}                                 & \multicolumn{3}{c|}{DIODE-Indoor}                                                                                         & \multicolumn{3}{c}{DIML-Indoor}                                                                                          \\
                                                     & $\delta_{1} \uparrow$ & $REL \downarrow$ & $RMSE \downarrow$ & \multicolumn{1}{l}{$\delta_{1} \uparrow$} & \multicolumn{1}{l}{$REL \downarrow$} & \multicolumn{1}{l|}{$RMSE \downarrow$} & \multicolumn{1}{l}{$\delta_{1} \uparrow$} & \multicolumn{1}{l}{$REL \downarrow$} & \multicolumn{1}{l}{$RMSE \downarrow$} \\ \hline
100\% : 0\% & 0.422                 & 0.229            & 0.939             & 0.226                                     & 0.368                                & 1.709                                  & 0.665                                     & 0.171                                & 0.516                                 \\

\textbf{60\% : 40\%}                                        & \textbf{0.787}        & \textbf{0.145}   & \textbf{0.588}    & \textbf{0.522}                            & \textbf{0.283}                       & \textbf{1.363}                         & \textbf{0.796}                            & \textbf{0.147}                       & \textbf{0.414}                        \\
0\% : 100\% & 0.608                 & 0.182            & 0.764             & 0.312                                     & 0.339                                & 1.614                                  & 0.498                                     & 0.205                                & 0.619                                 \\
\hline
\end{tabular}}
\end{table*}

\subsection{Accuracy on original NYUDv2 (seen)}
Our model aims to improve the generalization ability of monocular depth estimation in unseen datasets without reducing its accuracy in seen datasets. Therefore, we test our model on the test examples of original NYUDv2. Notably the training and testing examples on original NYUDv2 are captured by the camera of the same focal length in similar scenes. Table \ref{nyu_o} concludes the results compared with not only the five recent baselines but also more other ones. Our model achieves the same or better results compared with all the baselines. For example, our model slightly outperforms the second-best model \cite{2} by 0.7\% in RMSE. It indicates that incorporating focal length does not reduce the performance of our model on the original NYUDv2.

\subsection{Examples in real scenes}
\begin{figure}[h]
  \centering
  \includegraphics[width=\linewidth]{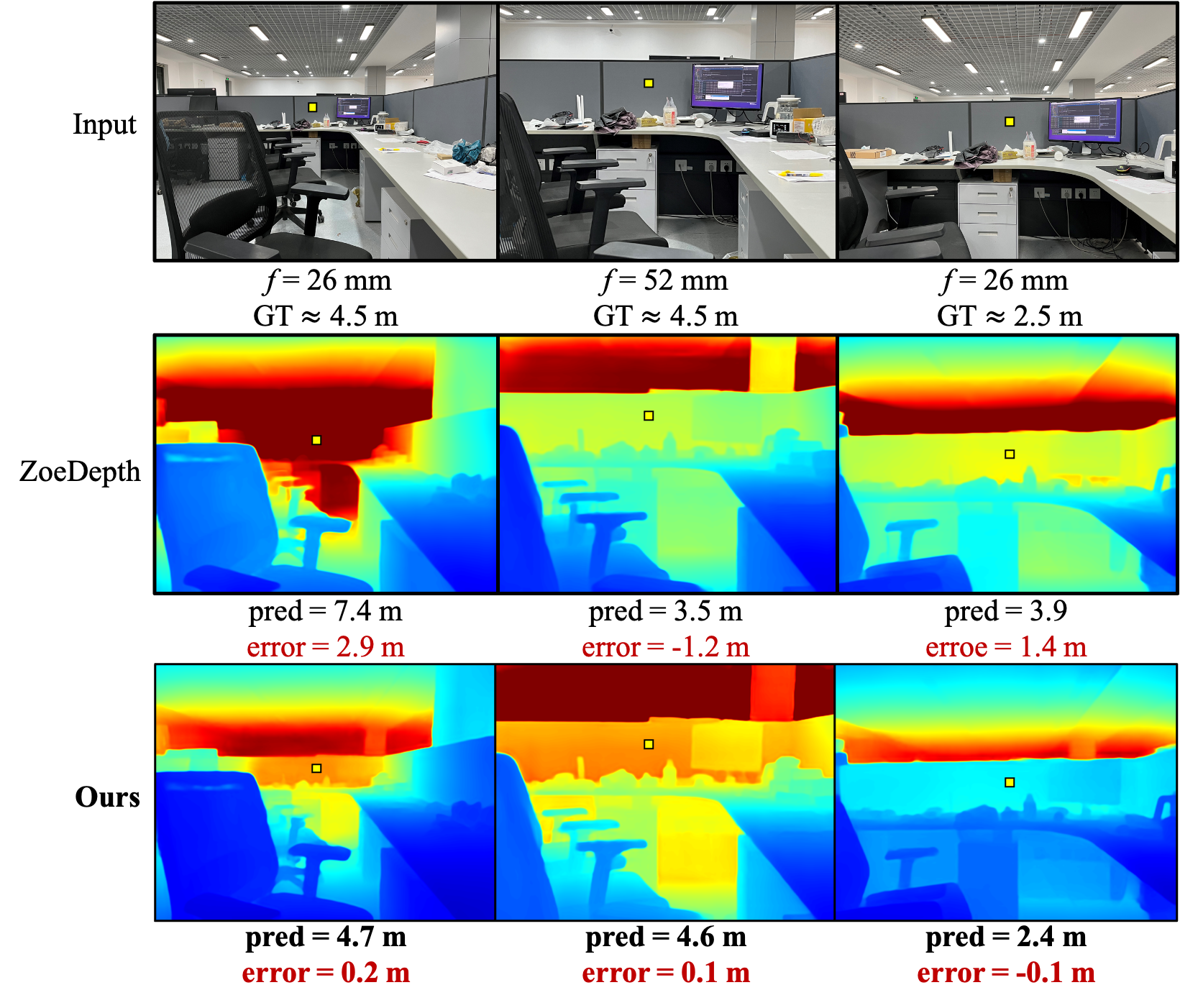}
  \caption{Examples in real scene. Our model accurately predicts the absolute depth values (\textit{i.e.} distances in meters) of the dot points compared with the baseline model \cite{2}.}  \label{real}
  \Description{A woman and a girl in white dresses sit in an open car.}
\end{figure}

We additionally test our model in real scenes. Figure \ref{real} gives three test images captured by a mobile phone with different focal lengths at different locations in our lab. We use our FS-Depth model and the second-best model \cite{2} to predict absolute depth maps from these images. The results show that the baseline model wrongly estimates the absolute depth values (\textit{i.e.} distances in meters), though structure of the scene is well predicted. By comparison, our model predicts the absolute depth values of the scene more accurately. It ensures that our model may be well implemented in practical scenarios.

\subsection{Ablation studies}

\noindent $\mathbf{Focal \ length \ incorporation.}$
In \textbf{Section 3.4}, we propose multi-scale features to incorporate focal length in multiple layers of the network, denoted “multi-scale”. We verify its effectiveness by comparing with two other ways: (1) our base model ZoeDepth \cite{2} that does not incorporate focal length in the network, denoted “base model”; (2) incorporating focal length in a single layer of our base model following the work \cite{1}, denoted “single-scale”. Notably the number of nodes in the last fully connected layer is adjusted to 768 to better fuse with the features of the base model. Table \ref{cmp} concludes the results of the ablation study. Our final model consistently achieves the best results. It indicates that focal length is considered more comprehensively in our final model.

\noindent $\mathbf{Learning \ rates.}$
In\textbf{ Section 3.6}, we determine the learning rate of the relative depth estimation network to be $1/50$ of the focal length features generation network and the absolute depth estimation network, denoted “$1/50 \times 1.6^{-4} $”. We verify its effectiveness by comparing with two general strategies: (1) freezing the learning rate of the relative depth estimation network, denoted “freeze”; (2) setting the three learning rates the same, denoted “$1.6^{-4}$”. Table \ref{lr} concludes the results of the ablation study. We observe that the learning rate has a significant impact on our model. The learning rate of our final model “$1/50 \times 1.6^{-4}$” achieves the best results with all metrics. Because the relative depth estimation network does not involve the input of focal length, a higher learning rate may introduce errors caused by the focal length to the relative depth estimation network. Simply freezing the weights of the relative depth estimation network would lead to excessive learning burden, hindering the performance of the absolute depth estimation network. The learning rate in our final model makes a balance between the two cases above.

\noindent $\mathbf{Data \ augmentation \ ratio.}$
In\textbf{ Section 4}, the RGB-Depth pairs in Step 2 and Step 3 are mixed in a ratio of 60\%:40\% to generate an augmented training dataset with diverse focal lengths. We verify its effectiveness by comparing with two other cases: (1) only Step 2 is activated, denoted “100\%:0\%”, (2) only Step 3 is activated, denoted “0\%:100\%”. Table \ref{ratio} concludes the ablation studies on three original unseen datasets. 
We observe that the ratio has a significant impact on the generalization ability of monocular depth estimation. 
The ratio of our final model “60\%:40\%” consistently achieves the best results with the metrics in all test datasets. In the case of “100\%:0\%”, the network may be overfitting to a single depth range and fail to well learn the meaning of input focal lengths. In the case of “0\%:100\%”, some prior knowledge such as the size of objects may be destroyed and monocular depth estimation could not well learn such prior knowledge. The data augmentation ratio in our final model makes a balance between the two cases above.



\begin{table}[]
  \caption{Accuracy on original NYUDv2 (seen). Incorporating focal length does not reduce the performance of our model on the original NYUDv2.}
  \label{nyu_o}
\setlength{\tabcolsep}{1.5mm}{
\begin{tabular}{c|cccccc}
\hline
Method                & $\delta_{1} \uparrow$ & $\delta_{2} \uparrow$ & $\delta_{3} \uparrow$ & $REL \downarrow$ & $RMSE \downarrow$ & $log_{10} \downarrow$ \\ \hline
Eigen \textit{et al.} \cite{39} & 0.769                 & 0.950                 & 0.988                 & 0.158            & 0.641             & \_\_\_\_              \\
Laina \textit{et al.} \cite{40} & 0.811                 & 0.953                 & 0.988                 & 0.127            & 0.573             & 0.055                 \\
Hao \textit{et al.} \cite{41}  & 0.841                 & 0.966                 & 0.911                 & 0.127            & 0.555             & 0.053                 \\
DORN \cite{22}                 & 0.828                 & 0.965                 & 0.992                 & 0.115            & 0.509             & 0.051                 \\
SharpNet \cite{42}             & 0.836                 & 0.966                 & 0.993                 & 0.139            & 0.502             & 0.047                 \\
BTS \cite{38}                  & 0.885                 & 0.978                 & 0.994                 & 0.110            & 0.392             & 0.047                 \\
Yin \textit{et al.}   & 0.875                 & 0.976                 & 0.994                 & 0.108            & 0.416             & 0.048                 \\
AdaBins \cite{6}              & 0.903                 & 0.984                 & 0.997                 & 0.103            & 0.364             & 0.044                 \\
LocalBins \cite{7}            & 0.907                 & 0.987                 & \underline{0.998}                 & 0.099            & 0.357             & 0.042                 \\
NeWCRFs  \cite{10}             & 0.922                 & \underline{0.992}                 & \underline{0.998}                 & \underline{0.095}            & 0.334             & \underline{0.041}                 \\
ZoeDepth \cite{2}             & \underline{0.955}                 & \textbf{0.995}        & \textbf{0.999}        & \textbf{0.075}   & \underline{0.270}             & 0.032                 \\ \hline
\textbf{Ours}         & \textbf{0.956}        & \textbf{0.995}        & \textbf{0.999}        & \textbf{0.075}   & \textbf{0.268}    & \textbf{0.032}       \\ \hline
\end{tabular}}
\end{table}

\begin{table}[]
  \caption{Ablation study of the learning rate in relative depth estimation (original NYUDv2).}
  \label{lr}
  \setlength{\tabcolsep}{1.5mm}{
\begin{tabular}{c|cccccc}
\hline
Method     & $\delta_{1} \uparrow$ & $\delta_{2} \uparrow$ & $\delta_{3} \uparrow$ & $REL \downarrow$ & $RMSE \downarrow$ & $log_{10} \downarrow$ \\ \hline
$1.6^{-4}$          & 0.948                 & 0.995                 & 0.998                 & 0.080            & 0.284             & 0.034                 \\
\textbf{$1/50 \times 1.6^{-4}$} & \textbf{0.956}        & \textbf{0.995}        & \textbf{0.999}        & \textbf{0.075}   & \textbf{0.268}    & \textbf{0.033}        \\
freeze          & 0.861                 & 0.984                 & 0.997                 & 0.121            & 0.437             & 0.053                 \\ \hline
\end{tabular}}
\end{table}

\section{Conclusion}
In this paper, we investigated the ill-posed problem of monocular depth estimation in unseen indoor scenes. We developed a focal-and-scale depth estimation model to well learn absolute depth from single images with diverse focal lengths and scene scales. Our model considerably improved the generalization ability of monocular depth estimation on unseen datasets. Notably, our model additionally incorporated focal lengths in monocular depth estimation while focal lengths cannot be directly fed into existing models. Our model provided a possible way to well predict absolute depth values in practical scenarios. In the future, we aim to well improve its generalization ability across indoor and outdoor scenes.

\clearpage
\bibliographystyle{ACM-Reference-Format}
\bibliography{sample-base}

\end{document}